\documentclass[letterpaper, 10 pt, conference]{ieeeconf}  

\IEEEoverridecommandlockouts                              

\usepackage[nolist]{acronym} 
\usepackage{svg}
\usepackage{layouts}
\usepackage{hyperref}
\usepackage{float}
\usepackage[T1]{fontenc}

\title{\LARGE \bf
How to Drive -  An Ability-based Description of Autonomous, Remote and Human Driving
}

\author{Florian Pfab, Nils Gehrke and Frank Diermeyer
\thanks{*This work was supported by the Federal Ministry of Education and Research of Germany~(BMBF) within the project ConnRAD~(FKZ~16KISR034), the project Wies'n Shuttle~(FKZ~03ZU1105AA) in the MCube cluster and through basic research funds from the Institute for Automotive Technology.}
\thanks{The authors are with the Institute of Automotive Technology at the Technical University of Munich (TUM), 85748 Garching bei Munchen,
Germany. Email: {firstname}.{lastname}@tum.de}%
}

\begin{document}

\maketitle
\thispagestyle{empty}
\pagestyle{empty}

\begin{abstract}
The development of autonomous and remote-operated driving systems requires extensive stakeholder analyses, requirement engineering, and formalized system descriptions. This is necessary to guarantee the success of the final product after the expensive and time-consuming development phase. To integrate a formalized description of the required abilites of the system, ability graphs have been proposed in the literature. Up to this date, however, this ability graph has only been used to model less complicated driver assistance systems in the literature. This work aims to introduce the value of an ability graph-based description of complex driving systems. This is achieved by successfully demonstrating and discussing a method for constructing a holistic ability graph capable of describing the entirety of abilities required for any driving system.
\end{abstract}
\section{INTRODUCTION}
\label{sec:introduction}

Autonomous Driving has the potential to transform transportation and society at large. 
It is estimated that the \ac{av} industry could create up to \$400 billion in revenue by 2035 \cite{mckinsey2023}. 
Recent implementations promise that the major technical challenges of \acp{av} can be resolved in the foreseeable future and that a crucial contribution of \acp{av} to the safety of public roads can be expected \cite{kusano2023comparison}. 
Still, significant advances have to be made in the upcoming years to fully evolve their economic and social potential and make a meaningful contribution to future mobility. 
To date, no \ac{sds} is capable of handling every scenario without any human intervention. 
While this problem will not be solved in the next years, robust vehicles without any human inside can still become a reality.
One technology that serves as an enabler is teleoperation. 
By transmitting and presenting sensor data from the vehicle to a human remote operator, the strengths of humans can be incorporated into the operation of \acp{av} \cite{Maj2022}. 

\subsection{Potential Application Areas}
The major challenge for such mobility systems is ensuring safe operation in any given situation. 
A potentially crucial tool to achieving this is self-awareness of the \ac{sds} regarding its abilities \cite{Dutt2020}. To this day, there is no available literature providing a precise description of the necessary capabilities of a system participating in public traffic. 
This is majorly due to the high complexity of the required system and, thus, the resulting description.
Given a representation of the required abilities, multiple issues currently existing throughout the development and operation of remote or automated driving systems can be addressed.

A first issue is the required completeness of the developed software solution to enable legal operation on public roads.
Requirements for systems are usually defined at the beginning of the development phase.
This means that missing a specific requirement during this initial system development phase due to the overbearing overall complexity can potentially result in high code refactoring efforts and consequently high costs or even loss of public trust in the system \cite{StatementCruise}.
This risk can be mitigated by applying a methodically robust ability graph as a reference to compare the system requirements against for completeness.

Further, the definition of a holistic set of driving abilities enables developers during the testing phase to have an additional, system independent reference for the test design.
If required, the proposed graph can be further extended and refined to a lower level of detail by continuing the proposed method. This allows the mapping of software and hardware components at all levels to respective abilities on the same level of detail and thus to derive tests ranging from component tests to integration up to system tests via the holistic ability graph.

Lastly, the ability graph provides a framework for monitoring the overall system by measuring the impact of sub-abilities on the overall system performance. 
This can be in a first steps achieved by a binary evaluation, checking if all required abilities are currently covered by correctly running modules. 
An extended approach to the impact analysis could assign a continuous performance score to the individual sub-abilities.
This would allow for more in-depth monitoring of the system.

\subsection{Contributions}
With this publication, the authors present a description of the driving task that can be used for the above-mentioned applications.
A graph-based approach is presented to determine the necessary abilities. One major advantage of this approach is the solution-neutral design of it. The resulting graph is valid for any system design and can be used for the safety assessment of autonomous, remote, and human-controlled road vehicles.
\\
In section \ref{sec:relatedwork}, the current state of the art concerning graph-based capability representations is presented. In addition, literature presenting tasks necessary for vehicle guidance is reviewed. In \autoref{sec:abilitygraph}, the methodology used to generate one holistic ability graph comprising the sources from the preceding literature review is presented, and the resulting graph is illustrated. In \autoref{sec:discussion}, the obtained results are discussed, and the applicability of the graph is evaluated using different vehicle guidance systems. Section \ref{sec:conclusion} gives a summary and an outlook of future applications for the presented ability graph.

\section{RELATED WORK}
\label{sec:relatedwork}

This section presents previous work in the relevant research areas for this publication. It first introduces crucial contributions to using ability and skill graphs in autonomous vehicle systems. Subsequently, publications addressing tasks involved in vehicle guidance are presented.

\subsection{Ability and skill graphs for \acp{av}}

The initial proposal to use capabilities to model \ac{av} systems was made by Maurer \cite{Maurer2000}. Quality metrics are used to evaluate the system's capabilities to ensure appropriate behavior of a vision-based \ac{av}.

The work is further extended by Siedersberger \cite{Siedersberger2003}. He proposes that multiple capabilities needed to execute a specific task can be grouped into more complex, combined capabilities. With this tweak, hierarchical capability systems arise which can be depicted in a graph-based way.

Pellkofer \cite{Pellkofer2003} grouped these hierarchical capabilities into different categories (e.g., action primitives on the lowest level, schematic, combined capabilities in the middle level, and behavior capabilities on the top level), resulting in a directed, cyclic-free graph.
\cite{Bergmiller2011} introduces a skill network as part of a self-concept to enable monitoring and fault detection. 


 \begin{figure}
     \centering
     \includegraphics{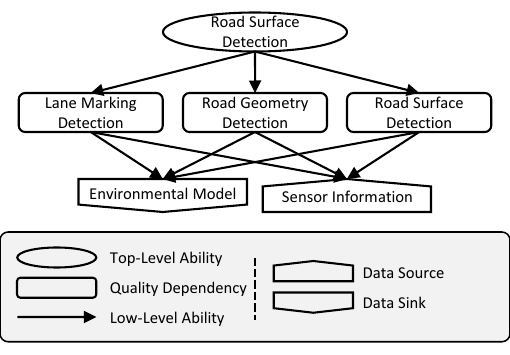}
     \caption{Examplary ability graph for the road surface detection task.}
     \label{fig:example}
 \end{figure}

  \begin{figure*}[]
     \centering
     \includegraphics{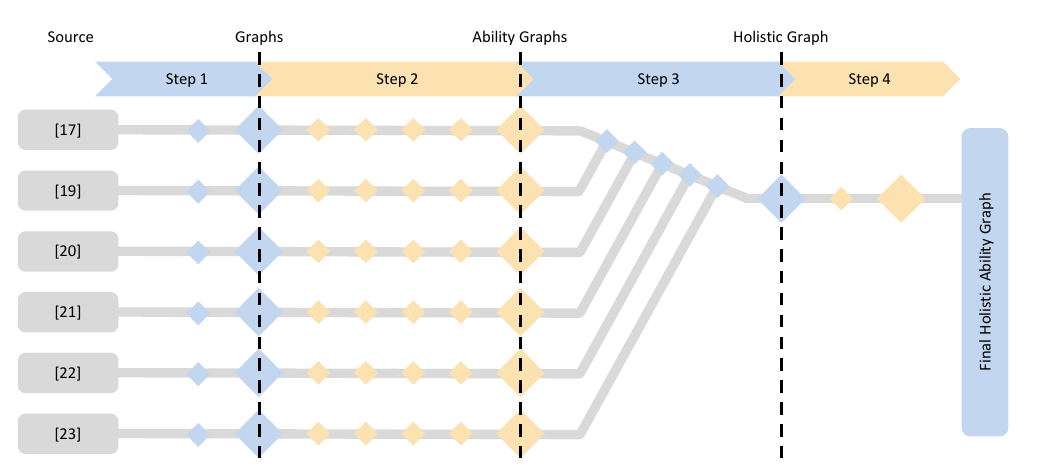}
     \caption{Representation of the method used for constructing the holistic ability graph.}
     \label{fig:our_method}
 \end{figure*}

Reschka \cite{Reschka2012} implements a safety system that performs degradation actions to keep the operation of the vehicle in a safe state. Specific performance criteria (e.g., grip value, position accuracy etc.) are identified, monitored, and according actions (e.g., modification of the maximum speed) are taken.

In \cite{Reschka2015}, the terms skill and ability are introduced to further specify the capabilities of a system.
While abilities are considered as the entity of conditions necessary to deliver a performance, skills describe an activity related to a certain task, including a performance (skill) level. \autoref{fig:example} illustrates an exemplary ability graph.

The usage of these graphs in the development process is further refined. 
The ability graph can be used in the item definition according to the ISO 26262 standard \cite{ISO26262}.
After the implementation of the system, the ability graph can be transferred to a skill graph, which can later be used for online monitoring of a driving system.

It is noted that the terms skill and ability are not always used consistently. This routes back to the translation from mostly German sources in the early stages of this research field. The authors use the taxonomy as presented in \cite{Reschka2015}.

\cite{Bagschik2016, Nolte2017, Reschka2017, Jat2021a} further apply the idea of ability and skill graphs on \ac{ad} systems. They have in common their deduction of ability graphs from specific scenarios, lacking a holistic depiction of the driving task.
This implies that specific scenarios need to be defined prior to the implementation of a system. In future \ac{ad} implementations heading towards SAE Level 5 \cite{SAEj3016}, this can not be assumed. An ability graph depicting all abilities needed for safe operation of vehicles on public roads is currently not available. With this publication, the authors aim to close this gap.

\subsection{Vehicle driving tasks}



Foundation of the holistic ability graphs are driving task descriptions and driver ability descriptions from literature. The research community has tackled the field of driving tasks, especially for \ac{ad} systems and human drivers. This initial literature review aims to identify relevant descriptions from all fields of application, including human drivers, teleoperated vehicles, and \acp{av}. This is particularly important when describing the required abilities indifferent to the driving solution in detail. A concise summary is given in \cite{Winner2015}.

\subsubsection{Driver-vehicle-environment system model}
This model, first described by Wickens \cite{Wickens1992}, explains human information processing with a model based on the processing stages 
considering the limitation of available resources.
The model takes the vehicle guidance task, the route, and environmental influences 
into consideration and provides mobility, safety, and comfort.

\subsubsection{Driving task allocation}
Bubb \cite{Bubb2003} establishes a categorization into primary, secondary, and tertiary driving tasks.

\subsubsection{Requirements from the vehicle guidance task}
Based on a questionnaire
, requirements are identified that need to be fulfilled for vehicle guidance \cite{Fastenmeier1995}. The requirements can be grouped into four categories, information sources, assessment tasks, decision and thinking processes, and vehicle handling.


\subsubsection{\ac{ddt}}
The \ac{sae} defines the \ac{ddt} as all real-time operational and tactical functions required to operate a vehicle in on-road traffic, excluding strategic functions \cite{SAEj3016}. Six subtasks (e.g., lateral vehicle motion control, etc.) are considered to be part of the \ac{ddt}.


\subsubsection{Three-level hierarchy of the driving task}
Donges \cite{Donges1982} models the driving task using the three levels navigation, guidance and stabilization.


\subsubsection{AV}
Pendleton \cite{Pendleton.2017} gives a concise overview of the components needed for \ac{av} software. The classification into perception, planning, control and vehicle cooperation has been established in research in recent years.

\section{\uppercase{The Ability-Graph Description for Driving}}
\label{sec:abilitygraph}

The ability graph aims to present a structured description of driving tasks independent of the \ac{odd}, the concrete realization of the driving system, and the driving task modality.
This includes all potential driving scenarios in adverse weather conditions, both human and machine drivers located in the vehicle or at a remote facility. 
During the process of constructing the ability graph, the focus was put on vehicle guidance on public roads.
This premise is reflected in the final graph's later discussion.

Commonly, the description of the required ability and the split of abilities into their sub-abilities is performed during the development process according to \cite{Nolte2017}.
This approach can not be applied for the targeted holistic ability graph, as such a development process would be too complex. Due to its complexity, it further presents the potential to miss important sub-abilities during the process.

With holistic being a strong requirement, the authors followed a strict method during the graph generation, which will presented in the following in more detail.
To achieve the subject, the authors propose a method using existing driving task descriptions from different domains and transforming them into one holistic graph.
Via this approach, the potential miss of sub-abilities in the existing literature due to the individual focuses of the driving task descriptions can be compensated while still using well-established and discussed task descriptions.
A crucial point in the method remains a profound discussion of the obtained graph to ensure its holistic nature.
 
The applied method consists of four consecutive steps (\autoref{fig:our_method}).
The method starts with the conversion of the selected driving task descriptions into a common, graph-based format. 
Common representations of driving tasks vary in the form of presentation. Thus, the second and third steps process the available representation to derive representations that can be compared against each other and combined into one holistic ability graph.
In the second step, the graphs are transformed into weakened ability graphs that fulfill the majority of the ability graph definitions.
A third step comprises of the merge of the individual, weakened ability graphs into one holistic ability graph.
The resulting ability graph contains a high amount of edges and nodes with similar but not equivalent sub-abilities.
To ensure the usability of the holistic graph, a fourth step is added to reduce the graph complexity while retaining all included sub-abilities.

\subsection{Graph deduction of existing task descriptions (Step 1)}
The selected  sources of driving task descriptions for the ability graph construction are \cite{Bubb2003, SAEj3016, Pendleton.2017, Fastenmeier1995, Donges1982, Wickens1992}.
The original representation of the driving tasks ranges from textual descriptions \cite{Bubb2003, SAEj3016, Pendleton.2017}, graphical representation of information flows \cite{Wickens1992, Donges1982} to hierarchical enumerations \cite{Fastenmeier1995}.
Based on these different information types, according strategies to leverage the represented information into a common graph representation are defined and applied.

Graphical representations can be directly transformed into graphs by defining nodes according to element definitions in the figures and edges according to the connections between the elements.
In \cite{Wickens1992}, some connections are between elements that comprise themselves.
The derived graph representation strictly follows the edge connection according to the sources and does not connect the sub-ordinary elements.

The hierarchical enumeration \cite{Fastenmeier1995} is transformed into a graph by representing each level as a new set of sub-abilities to the superior level. 

The extraction from textual representations \cite{Bubb2003, SAEj3016, Pendleton.2017} requires extracting abilities within the text and then linking the identified abilities via edges in the graph according to the contextual linkage within the text.
This includes dependencies on a function level, a logical level, and an information level.
This process required detailed reading and discussion by the authors.

\subsection{Transformation into ability graphs (Step 2)}
Based on the available graph representations of the previous step, this step comprises transforming the graphs into weakened ability graphs.
The ability graph definition according to \cite{Reschka2015} is used.
Ability nodes must be formulated as abilities, solution-neutral, and have either none or at least two sub-abilities.
Edges represent the quality dependency of abilities. Leaves of the ability graph must always be data sinks or sources. 
To not loose data during the merge, this definition is weakened to include abilities with one sub-ability during this step.\\
It is important to note that the edges of the obtained ability graph do not represent the information flow. 
Given the example of reading a book split into the ability to translate written letters to words and the ability to interpret a set of words as sentences, both abilities would not be dependent on each other. 
The main ability of reading a book highly depends on both abilities, the second ability however only requires a set of words to perform well without the necessity to be the correct set of words from the page. 
From an information propagation point of view, however, the second ability strongly depends on the ability to obtain words for a successful reading of a book.
\\
To transform a given graph into an ability graph, first, the given nodes are renamed to abilities. 
An exemplary renaming is the transformation of the node \textit{road network} to an ability \textit{perceiving the road network}.
Extensive work with the original source is required at this stage to identify and formulate the appropriate ability for each node.\\
In a second step, these nodes are transformed into solution-neutral formulations or removed if this can not be performed successfully.
Solution-neutral is not given if a domain-specific mean of realization is given.
Domains are considered to  be a human driver, a \ac{sds}, or systems using both software and human input. 
For example, listening to the audio output of a navigation system may be a relevant ability for a human driver, but it is primarily focused on human users. 
A more generalized description, allowing for a solution-neutral formulation, would be to obtain navigation information.
This step thus focuses on removing any predefined, domain-specific interfaces from the node formulation.\\
In a third step, existing edges are reevaluated. 
In case of only information flow depicted by the edges, these edges are removed.
If not present, additional data sources and sinks are added to substitute nodes from leave positions.
Additional edges are added to represent missing quality dependencies if necessary. 
Ability nodes with one sub-ability are concatenated to one single ability, and data sinks and sources are added accordingly.\\
After these steps, the six graphs are transformed into six ability graphs.
Due to the criticality of preserving the driving task descriptions from the original sources throughout this transformation process, the described steps were again performed within the group of the authors and reviewed for each graph modification. 

\subsection{Merge into a hollistic graph (Step 3)} \label{subsec:merge_holistic_graph}
In this step, the ability graphs are merged into the holistic graph for driving tasks. 
This is performed by identifying identical nodes in the six individual ability graphs. Based on these identical nodes, a merge can be performed. 
Identical nodes consist of nodes describing identical abilities.
This means, for example, that the abilities \textit{perception}, \textit{environmental perception}, \textit{monitoring the driving environment}, \textit{perceiving the environment} and \textit{retrieving information, sensing and perception} are considered identical and merged. 

\begin{figure}
    \centering
     \includegraphics[width=\linewidth]{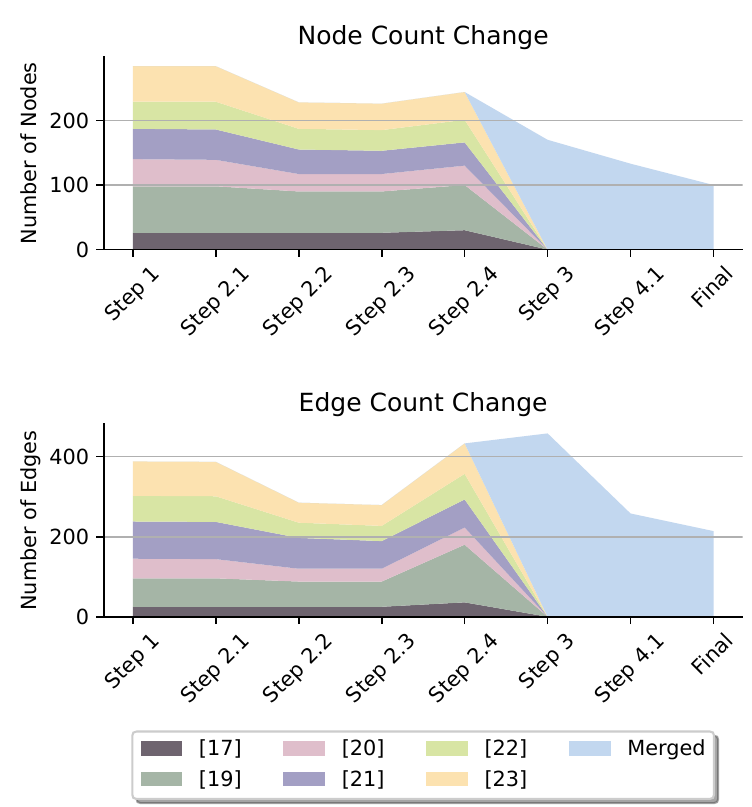}
    \caption{The node and edge count change during the different stages from the original driving task literature towards the final holistic ability graph as presented in \autoref{sec:abilitygraph}.}
    \label{fig:graphsize}
\end{figure}

\begin{figure*}
    \centering
     \includegraphics{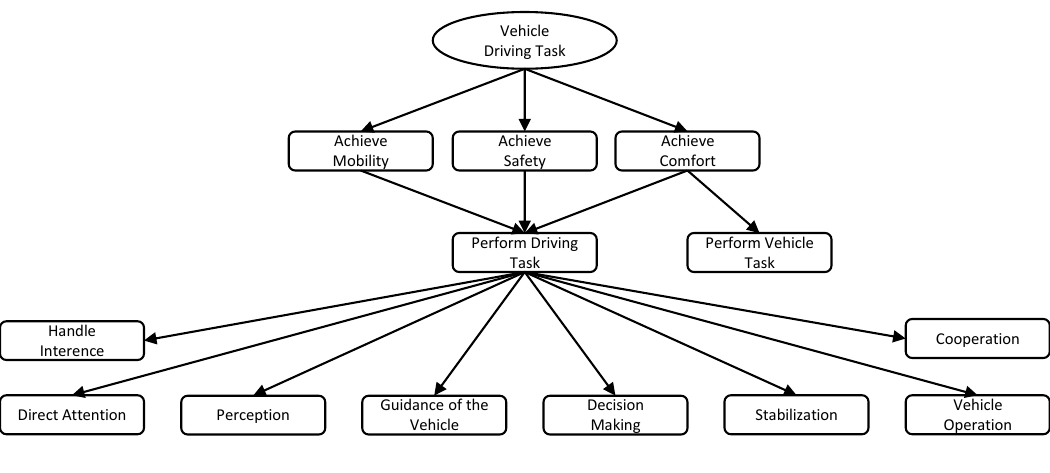}
    \caption{The final holistic ability graph. To reduce the complexity in this figure, the level of sub-abilities is reduced to four. More detail can be obtained by visualizing the graph using the tool in \cite{LinkGraph}.}
    \label{fig:finalabilitygraph}
\end{figure*}

Other abilities, counting localization to perception, are not considered identical.
All authors reviewed and approved all decisions on identical nodes.\\
To make sure to compare all abilities in all graphs, starting at one ability graph, all other ability graphs are mapped to the base ability graph one by one.
Identical nodes are then replaced in the currently considered ability graph. 
The edges remain untouched in this process. 
Thus, the obtained merged graph may contain more than one edge connecting the same two nodes.
\autoref{fig:graphsize} shows the changes in the number of nodes and edges in the different stages towards the holistic ability graph.

\subsection{Enhancing the usability (Step 4)}
To enhance usability and facilitate the later discussion, the holistic ability graph is reduced in complexity to a broader level of detail while maintaining the original relations.
This is achieved by clustering certain abilities identified as similar but not identical in section \ref{subsec:merge_holistic_graph} and by clustering very high detail sub-abilities. 
The resulting graph is proposed to be viewed without data sinks and sources for better readability.
This possibility is also provided by the tool published by the authors containing the final holistic ability graph and allowing reproducing the development of the graphs through the multiple steps \cite{LinkGraph}.
The first four levels of the final ability graph are represented in \autoref{fig:finalabilitygraph}.


\section{\uppercase{Discussion}}
\label{sec:discussion}
This section aims to identify the limits in both the level of detail of the ability graph and its holistic nature considering potential applications.
Therefore, different application use cases are required.

\begin{figure}
     \centering
     \includegraphics{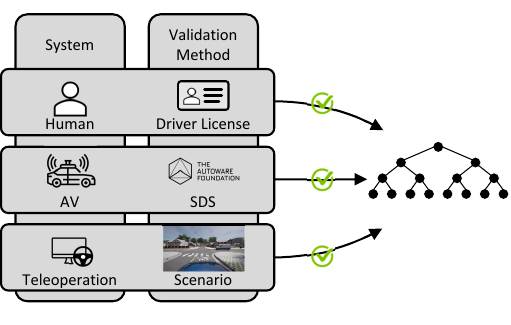}
     \caption{The different applications and methods used to validate the derived ability graph.}
     \label{fig:discussion}
 \end{figure}

A reference for the successful deployment of driving systems is the human driver. 
While describing the overall driving tasks performed by a human driver would effectively substitute the approach pursued in this work and thus not be suitable, the learning process of human drivers that culminates into the driver's license exam can be used.
Given the holistic nature of the final ability graph, this set of sub-abilities must be sufficient to pass the exam.
The driving exam in Germany defines multiple requirements \cite{drivingexam}.
Further references for driving task descriptions are taken from a remote operation scenario \cite{Brecht2024} and an open source software stack \cite{AutowareUniverse} that aims to fully cover the autonomous driving pipeline.
By comparing these implementations with the graph, both limits in the graph's level of detail and the question of how to correctly apply and use the final graphs are addressed. \autoref{fig:discussion} visualizes the validation methods used for the different applications.

\subsection{Can the graph pass a human driver's license test?}
A mandatory task during the exam is a reverse parking maneuver or turning the vehicle around using a combination of reverse and forward driving \cite{drivingexam}.
According to \cite{drivingcatalogue}, this task comprises a set of tasks that must be successfully completed.
The student needs to monitor the traffic, position the vehicle appropriately, adjust the speed accordingly, communicate with the environment, and demonstrate his or her environment awareness.
In detail, monitoring the environment comprises checking for obstacles within the dead angle.
Positioning the vehicle requires the student to stay on the right side of the road while performing the reverse maneuver, not blocking the street. 
Vehicle operation means that the student is able to turn around using a maximum of two correction moves.
This task is compared to the final ability graph for discussion.

Monitoring the perception, including the dead angle, is given by the general description of the perception ability in the holistic graph.
The velocity adjustment and the communication are given via the \textit{Vehicle Operation} ability and its sub-abilities, including the option to select the reverse gear of the vehicle. Therefore, the requirements for the driving exam are met.

Vehicle operation and positioning are not directly stated in the ability graph.
In the case of vehicle operation, the requirement from the driving exam highly depends on the implementation quality of the \textit{Guidance of the Vehicle} ability.
In addition, the road geometry can be perceived using the \textit{Perception} ability.
However, the required driving on the right sight is not directly part of the ability. When looking closely at the graph's creation, it becomes clear that this is part of obeying traffic rules, which are part of the graph.
Thus, to correctly apply the holistic ability graph, this example shows that also the derivation of the graph needs to be considered. 

\subsection{Does the graph contain every module within an open-source \ac{av} stack?}
The Autoware foundation maintains a modular open-source autonomous driving software\cite{AutowareUniverse}.
The software is structured into different modules that fulfill different skills of the \ac{ad} software \cite{AutowareNodes}.
With the goal of achieving a fully \ac{ad} stack capable of driving on public roads, the modular approach is best suited to be compared against the final ability graph for the discussion.
In this discussion, only the highest level of representation in the given module overview \cite{AutowareNodes} is considered and compared against the ability graph.\\
Starting with sensing, the considered software stack contains multiple modules for camera, radar, lidar, and GNSS sensing. 
These sensing pipelines aim to capture and preprocess the incoming sensor data for later detection and prediction tasks. 
In the ability graph, the raw sensor information is represented by the \textit{Information from Sensors} data source.
Preprocessing for the individual tasks is not directly part of the ability graph.
Rather, these steps can be considered part of the individual sub-abilities using the stated sensor source.
The level of detail could be refined by adding additional literature here if required.
The difficulty with an increase in detail is to keep the solution neutral form of the ability graph.\\
Further steps in the software modules include processing the sensor data to obtain a list of objects around the vehicle, detecting and classifying traffic lights and crosswalks, and creating an occupancy grid. These results are then used in the planner together with a current pose to define behavioral goals and plan a trajectory based on a previously generated mission. The resulting trajectory is executed via a trajectory follower and a lower-level vehicle interface.
These modules can be mapped to abilities of the final graph. 
Interestingly, while the perception of traffic signs is specifically listed as a sub-ability of \textit{Perception}, the perception of traffic lights is only summarized under the example \textit{event-based monitoring} of the ability \textit{Environmental Perception}.
This shows that the usage of six sources for the construction of the graph and the transformation processes applied, result in a different level of detail in the ability abstractions.\\
Further, it can be noted that some abilities of the graph, including road surface state, weather or traffic signs, and perception of acoustic information from the final ability graph, are not represented in the considered software stack.
In addition, some abilities (e.g. \textit{Perform Vehicle Task}) are not present in \cite{AutowareNodes}.
The selected level of detail for the visualization \cite{AutowareNodes} causes the absence of some existing software implementations in the module graph, but the absence of abilities is mainly caused by missing functionalities of the software stack.
It can, however, be observed that the constructed holistic ability graph exceeds the current modular representation of the software stack, which supports the assumption of its completeness.

\subsection{Can the graph solve a teleoperation (\ac{av} disengagement) scenario?}
\cite{Brecht2024} provides multiple teleoperation scenarios that properly depict \ac{av} disengagements. To demonstrate the capability of the graph to solve realistic teleoperation scenarios, the ability graph is tested with one of these scenarios. The selected scenario 3 involves a failure of the planning module due to an incorrectly parked vehicle at the intersection's entry. The \ac{av} stops with the intention of giving the right of way to the parked vehicle. To add another level of complexity, bad weather conditions from scenario 1 are assumed.

The perception abilities given in the presented graph enable a system encountering this scenario to detect the parked vehicle at the intersection and its right of way coming from its position in relation to the ego vehicle. To cope with the bad weather conditions, it is necessary to perform secondary driving tasks (e.g., enhancing conspicuity, controlling the windscreen wiper, etc.). In addition, the surface of the road needs to be perceived to properly estimate a safe driving velocity. When assuming fully pronounced abilities of the vehicle, the scenario can be solved because the prediction ability allows assumptions about the future behavior of the parked vehicle, and the full knowledge about the applicable traffic rules allows the ego vehicle to continue its journey.

This scenario shows that the ability graph presented can also handle scenarios considered of high importance for teleoperated road vehicles.

\section{\uppercase{Conclusion and Future Work}}
\label{sec:conclusion}

To enable safe \ac{ad} on level 5 it is crucial for a system to know its own abilities at any time. With our efforts, a methodology has been developed to create a holistic graph that shows the abilities needed to safely perform the driving task on public roads. A graph based on current literature is presented. One major advantage of the resulting graph is its applicability for any driving system, ranging from a human driver to a teleoperated vehicle to an \ac{av}. To show the capability of the graph to properly depict the abilities of current applications, it is validated using a scenario from a human driver license test, the modules of an \ac{sds}, and a use case for teleoperated road vehicles. To the authors' knowledge, the graph is exhaustive, although the presented methodology allows the future extension of the graph.
\\
The presented ability graph can be used to set requirements for any driving system. In the later development process, it can be transferred to a skill graph which can be used for online monitoring of the system and can thereby make a crucial contribution to the safety of the system. The presented ability graph strengthens the validity of such a safety system as it provides a way to ensure the completeness of the monitoring module. 
\\
Another important application can be expected by opposing the ability graph to the skill graph of a teleoperation system. By comparing the provided skills of the implementation of the system to the abilities needed for the safe operation of a road vehicle, a responsibility allocation between human operator and teleoperation system can be achieved.
\section*{\uppercase{Acknowledgments}}

Florian Pfab, as the first author, contributed the initial idea of the holistic ability graph. Nils Gehrke provided crucial input for the concept of the paper and for creating the final ability graph. Frank Diermeyer made essential contributions to the concept of the paper, revised the manuscript critically for important intellectual content, and gave final approval for the published version. He agrees with all aspects of the work.



\bibliographystyle{IEEEtran}
\bibliography{literature}

\begin{acronym}

    
    \acro{av}[AV]{Automated Vehicle}
    \acroplural{av}[AVs]{Automated Vehicles}
    \acro{ad}[AD]{Autonomous Driving}                
    
    \acro{odd}[ODD]{Operational Design Domain}
    \acro{sae}[SAE]{Society of Automotive Engineers}
    \acro{ddt}[DDT]{Dynamic Driving Task}{}
    \acro{sds}[SDS]{Self Driving System}{}
    \acroplural{sds}[SDSs]{Self Driving Systems}{}
    \acro{acc}[ACC]{Adaptive Cruise Control}

\end{acronym}
\newpage

\end{document}